\documentclass[11pt]{article}

\usepackage[preprint]{acl}

\usepackage{times}
\usepackage{latexsym}

\usepackage[T1]{fontenc}
\usepackage[utf8]{inputenc}

\usepackage{microtype}

\usepackage{inconsolata}

\usepackage{graphicx}

\usepackage{amsmath}    % 数学公式基础包
\usepackage{amsfonts}   % 提供\mathbb等符号
\usepackage{threeparttable}
\usepackage{booktabs}
\usepackage{bm}
\usepackage{multirow}
\usepackage{xcolor}          % 加载颜色包
\usepackage{colortbl}        % 表格着色核心包
\definecolor{gray-bg}{gray}{0.95}  % 0=纯黑，1=纯白，0.9=浅灰
\usepackage{arydshln} % 虚线专用

\usepackage{stfloats}
\usepackage{float}
\usepackage{enumitem}
\usepackage{hyperref}

\usepackage[linesnumbered,ruled,vlined]{algorithm2e}

\usepackage{wrapfig} % 包围图片
\usepackage{balance}

\title{Efficient Learned Data Compression via Dual-Stream Feature Decoupling}

\author{
    Huidong Ma$^{1,2}$, Xinyan Shi$^{1}$, Hui Sun$^{1}$, Xiaofei Yue$^{3}$, \\\textbf{Xiaoguang Liu}$^{1*}$, \textbf{Gang Wang}$^{1*}$, \textbf{Wentong Cai}$^2$ \\
    $^1$ College of Computer Science, TMCC, SysNet, DISSec, GTIISC, Nankai University \\
    $^2$ College of Computing and Data Science, Nanyang Technological University \\
    $^3$ Beijing Institute of Technology \\
    $^{*}$ Corresponding authors \\
    {\tt\small \{mahd, liuxg, wgzwp\}@nbjl.nankai.edu.cn}
}

\begin{document}
\maketitle

% \begin{abstract}
%     While Learned Data Compression (LDC) has demonstrated superior potential in compression ratio, balancing precise probability modeling with system efficiency remains challenging. This limitation stems from uniform single-stream architectures, which struggle to simultaneously capture micro-syntactic and macro-semantic features, forcing a reliance on deep serial stacking that exacerbates latency. Compounding this, heterogeneous systems are constrained by device speed mismatches and Amdahl's Law in serial processing, limiting throughput. To this end, we propose a Dual-Stream Multi-Scale Decoupler that disentangles local and global context to replace deep serial processing with shallow parallel streams, and incorporate a Hierarchical Gated Refiner for precise instance adaptation. Furthermore, we design a Concurrent Stream-Parallel Pipeline, which overcomes systemic bottlenecks to achieve full-pipeline parallelism. Extensive experiments demonstrate that our method achieves state-of-the-art performance in both compression ratio and throughput, while maintaining the lowest latency and memory usage. The code is available at \url{https://anonymous.4open.science/r/FADE-D817}.
% \end{abstract}

\begin{abstract}
While Learned Data Compression (LDC) has achieved superior compression ratios, balancing precise probability modeling with system efficiency remains challenging. Crucially, uniform single-stream architectures struggle to simultaneously capture micro-syntactic and macro-semantic features, necessitating deep serial stacking that exacerbates latency. Compounding this, heterogeneous systems are constrained by device speed mismatches, where throughput is capped by Amdahl's Law due to serial processing. To this end, we propose a Dual-Stream Multi-Scale Decoupler that disentangles local and global contexts to replace deep serial processing with shallow parallel streams, and incorporate a Hierarchical Gated Refiner for adaptive feature refinement and precise probability modeling. Furthermore, we design a Concurrent Stream-Parallel Pipeline, which overcomes systemic bottlenecks to achieve full-pipeline parallelism. Extensive experiments demonstrate that our method achieves state-of-the-art performance in both compression ratio and throughput, while maintaining the lowest latency and memory usage. The code is available at \url{https://github.com/huidong-ma/FADE}.
\end{abstract}

\section{Introduction}
With the rapid evolution of the Internet and AI-generated content technologies, multi-source data (spanning text, multimedia, and scientific sequences such as genomes and floating-point data) is experiencing explosive growth at a pace far surpassing Moore's Law~\cite{sun2024lrcb,sun2024pqsdc,sun2025genomics,NNLCB,sun2025pmklc}. This surge imposes tremendous pressure on data transmission bandwidth and storage infrastructure. Traditional lossless compression algorithms, represented by Gzip~\cite{gzip}, zstd~\cite{zstd}, and others~\cite{bzip2,deutsch1996deflate}, rely primarily on heuristic dictionary matching (e.g., LZ77~\cite{ziv1977universal}) or statistical coding (e.g., Huffman~\cite{huffman1952method}, ANS~\cite{duda2013asymmetric}). However, they struggle to effectively capture the high-order semantic redundancy in complex data, resulting in limited compression capability.

\begin{figure}[t]
    \centering
    \includegraphics[width=\columnwidth]{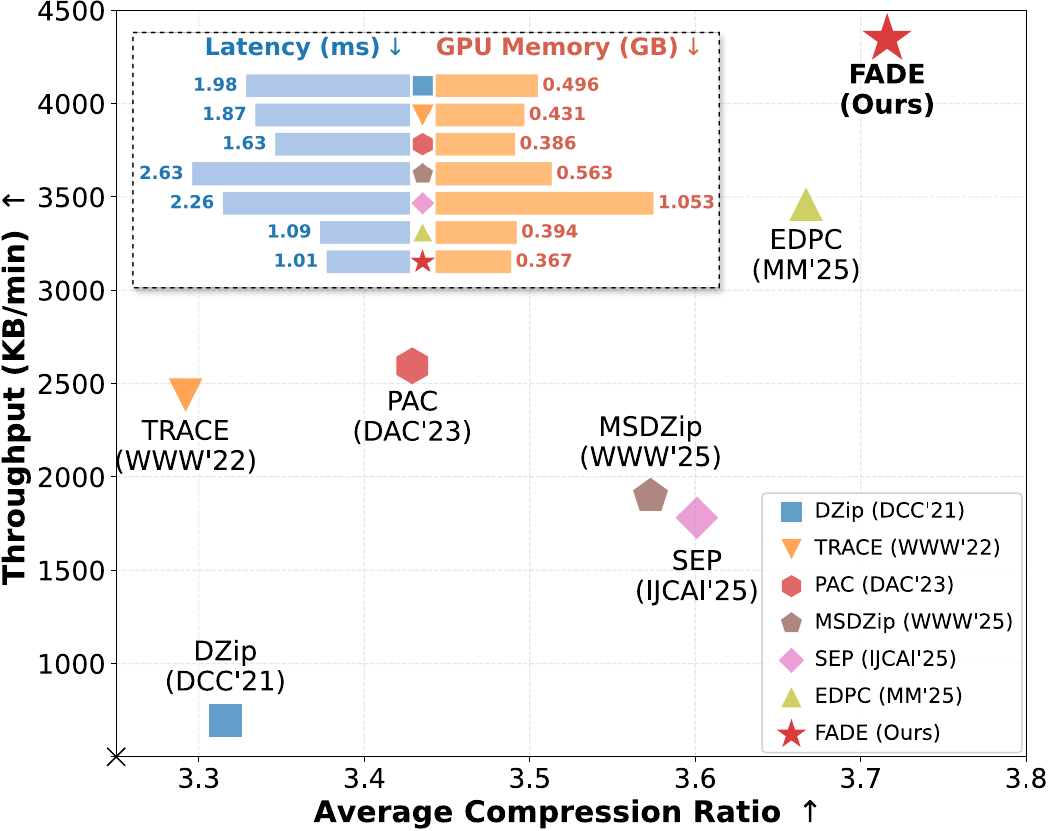}
    \caption{Trade-off between compression ratio and throughput. Top-right is better.}
    \label{fig:cr_vs_tp}
    \vspace{-0.38cm}
\end{figure}

Recently, deep learning has revolutionized sequence modeling, enabling LDC to significantly outperform traditional methods in compression ratio~\cite{NNLCB}. Despite this progress, balancing precise probability modeling with system-level efficiency remains challenging due to two structural limitations: \textbf{First}, uniform single-stream architectures struggle to capture heterogeneous micro-macro patterns using unified parameters. Consequently, existing methods rely on deep Multilayer Perceptron (MLP) stacking to approximate complex distributions, a strategy that inevitably lengthens computational paths and severely exacerbates autoregressive decoding latency. \textbf{Second}, heterogeneous systems suffer from systemic throughput bottlenecks. The inherent speed mismatch between GPU probability generation and CPU arithmetic coding causes pipeline stalls, while autoregressive serial decoding remains strictly bound by Amdahl’s Law~\cite{amdahl1967validity}, preventing parallel acceleration and restricting overall throughput.

% 我们提出了FADE，达到优秀压缩率的同时拥有优秀的吞吐量，同时保持了低的latency和GPU memory usage.

In this paper, we propose an e\textbf{\underline{f}}ficient le\textbf{\underline{a}}rned \textbf{\underline{d}}ata compr\textbf{\underline{e}}ssion method (\textbf{FADE}) that achieves superior compression ratios and high throughput, while maintaining low latency and GPU memory usage. 
Unlike existing methods, FADE reframes the modeling of complex dependencies by decoupling conventional deep serial processing into shallow parallel streams. Specifically, FADE employs a Dual-Stream Multi-Scale Decoupler (DMD) to disentangle features into a micro-syntactic Convolutional Neural Network (CNN)~\cite{lecun2002gradient,ma2023cnnlsv,ma2023ricme} branch and a macro-semantic MLP branch, and fuses these local and global features via the proposed Content-Adaptive Router. 
To ensure precise probability estimation, we incorporate a Hierarchical Gated Refiner (HGR) that leverages dynamic gating to inject stream-specific persistent memory for instance adaptation, while utilizing a high-capacity network to capture complex global dependencies.
Furthermore, to break autoregressive serial constraints, we design a Concurrent Stream-Parallel Pipeline (CSPP) that hybridizes data parallelism with thread-safe, double-buffered temporal parallelism. Our contributions are summarized as follows:

% 我先把整个故事给理一遍：
% 首先FADE包含三个模块：
% 1) Dual-Stream Multi-Scale Decoupler. 2) Hierarchical Gated Refiner. 3) Concurrent Stream-Parallel Pipeline
% 1) Dual-Stream Multi-Scale Decoupler: 传统的基于MLP的模型无法兼顾宏观和微观，迫使需要堆叠多层来提取。.我们提出 DMD，通过并行双流架构将特征解耦，分别利用 CNN 分支捕捉微观语法细节，以及 MLP 分支建立宏观语义依赖， 最后提出 Content-Adaptive Router 对双流特征进行动态融合。
% 2) Hierachical Gated Refiner: 我们提出HGR，通过双层架构对特征进行有粗到细的精炼。在粗粒度阶段，在粗粒度利用Batch Matrix Multiplication (BMM)构建实例级上下文，并引入Self-Gating机制去除噪声。在细粒度阶段，通过高容量的Gated投影实现精准的特征拟合。
% 3) Concurrent

\begin{itemize}[leftmargin=*, nosep]
    \item \textbf{Dual-Stream Multi-Scale Decoupler.} We propose the DMD, which decouples features into macro-semantics and micro-syntax, processes them concurrently via MLP and CNN branches, and fuses them using a content-adaptive router.

    \item \textbf{Hierarchical Gated Refiner.} We introduce the HGR, which performs coarse-to-fine refinement by constructing stream-aware context and achieving precise feature memorization and modeling to optimize the compression ratio.
    
    \item \textbf{Concurrent Stream-Parallel Pipeline.} We design the CSPP, which hybridizes data parallelism with thread-safe and double-buffered temporal parallelism, to achieve zero-wait processing and higher throughput.

    \item \textbf{SOTA Performance.} Extensive experiments on standard datasets demonstrate that FADE outperforms state-of-the-art methods in both compression ratio and throughput (see Figure~\ref{fig:cr_vs_tp}).

\end{itemize}

\section{Related Work}
LDC methods typically combine a probability prediction model and an entropy coding algorithm. While the primary focus of current research lies in constructing accurate and lightweight probability models, a few recent works have targeted pipeline optimization to enhance throughput.

\textbf{Neural Autoregressive Probability Models}. Early research mainly leveraged Recurrent Neural Networks (RNNs)~\cite{elman1990finding} and their variants to model sequential patterns. Specifically, LSTM-Compress~\cite{lstm-compress}, NNCP~\cite{NNCP}, DeepZip~\cite{DeepZip}, and DecMac~\cite{DecMac} all adopted Long Short-Term Memory (LSTM)~\cite{LSTM} as their prediction model to capture long-range dependencies. To balance efficiency and performance, DZip~\cite{DZip} proposed a semi-adaptive framework combining bootstrap and supporter models. Subsequently, the latest MSDLC~\cite{MSDLC} improved modeling capability by introducing xLSTM~\cite{xLSTM}. With technological evolution, methods based on Transformers~\cite{transformer} and Large Language Models (LLMs) have developed rapidly. NNCP v2~\cite{bellard2021nncp} achieved excellent performance through relative positional encoding. TRACE~\cite{TRACE} significantly reduced inference latency by introducing a linear attention mechanism SLiM~\cite{likhosherstov2021sub,choromanski2020rethinking}; LMIC~\cite{LMIC} and LLMZip~\cite{LLMZip} established new state-of-the-art compression ratios leveraging pre-trained models but face enormous computational overhead. Hybrid ensembles like CMIX~\cite{CMIX} achieve exceptional compression performance but are practically limited by excessive computational complexity.

\textbf{Lightweight Architectures and Feature Refinement}. To address the slow inference of deep networks, MLP-based lightweight compression architectures such as OREO~\cite{OREO} and PAC~\cite{PAC} have become a research hotspot, substantially boosting speed through masking and caching mechanisms.
Recent research has further explored MLP potential to enhance feature representation. MSDZip~\cite{MSDZip} designed a local-global-deep mixing block to stabilize cold-start training. SEP~\cite{wan2025sep} introduced a semantics enhancement module to capture complex intra-patch relationships. EDPC~\cite{lu2025edpc} proposed a dual-path framework and a latent transformation engine to enrich feature flow and reduce GPU memory usage.

\textbf{Parallelism and System Optimization}. Beyond model architecture, parallelism is key to improving throughput. In terms of data parallelism, MSDLC introduced a parallel expansion mapper for chunk-based data processing, while MSDZip proposed a stepwise-parallel strategy to accelerate large-scale data compression using multiple GPUs. Regarding pipeline optimization, SEP designed multi-stream pipelines, effectively masking I/O and transmission latencies. EDPC further decoupled probability prediction from arithmetic coding, realizing heterogeneous GPU-CPU parallelism.
% However, constrained by the autoregressive property, parallelization at the decompression stage remains an insufficiently resolved challenge in this field.

\begin{figure*}[t]
    \centering
    \includegraphics[width=\textwidth]{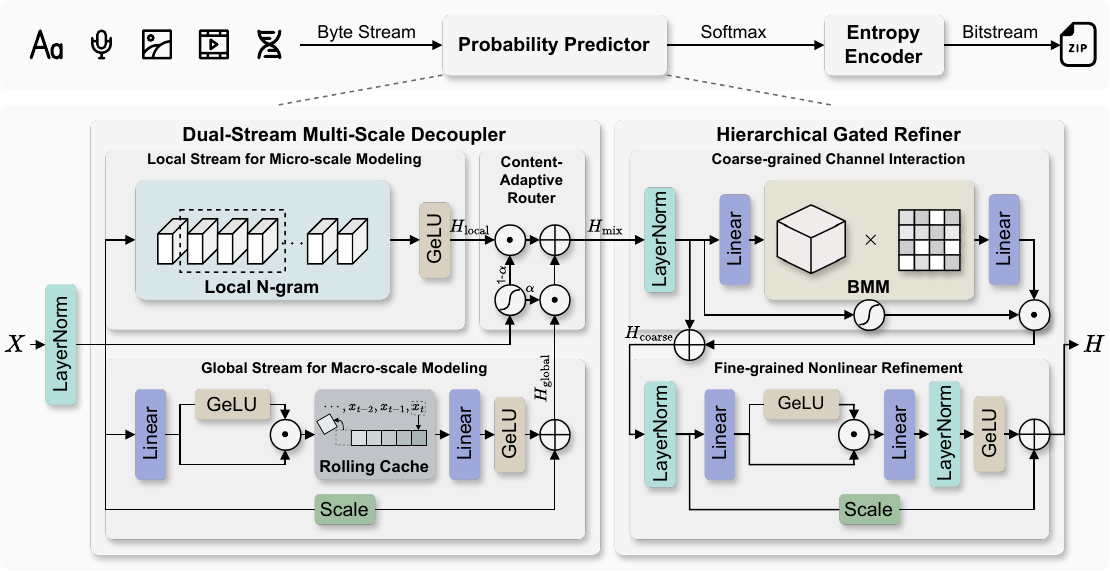}
    \vspace{-0.7cm}
    \caption{Overview of the proposed method. The embedded input $\bm{X}$ is disentangled into local and global contexts by the DMD, then fused and dynamically refined by the HGR to generate the final representation $\bm{H}$.}
    \label{fig:workflow}
    \vspace{-0.4cm}
\end{figure*}

\section{Method}
\subsection{Preliminaries}
\textbf{Problem Formulation.} 
LDC aims to map a discrete sequence of symbols $\bm{S} = \{x_1, \dots, x_n\}$ into the shortest bitstream. According to Shannon's source coding theorem, the expected coding length is lower-bounded by the entropy $H(\bm{S}) = - \sum P(\bm{S}) \log_2 P(\bm{S})$. Since the joint probability decomposes as $P(\bm{S}) = \prod_{t} P(x_t | x_{<t})$, approaching this theoretical limit relies on the accurate estimation of the conditional probability $P(x_t | x_{<t})$.\\
\textbf{Autoregressive Framework.}
As shown in Alg.~\ref{alg:compression}, to approximate this theoretical limit, LDC adopts an autoregressive framework including two phases:
\begin{itemize}[leftmargin=*, nosep]
    \item \textbf{Compression Phase.} At step $t$, the network $\mathcal{M}$ processes the history context $x_{<t}$ to predict the conditional distribution $\hat{p}_t$. An entropy encoder (e.g., Arithmetic Coding~\cite{witten1987arithmetic}) then utilizes this probability estimate to compress the target symbol $x_t$ into the bitstream.
    \item \textbf{Decompression Phase.} Operating as the inverse process while maintaining strict causality, the decoder employs the identical network $\mathcal{M}$ on the previously decoded context to reconstruct $\hat{p}_t$. Subsequently, the entropy coding algorithm recovers $x_t$ from the bitstream and appends it to the history for the next iteration.
\end{itemize}

% 基于此基础自回归框架，我们提出了xxxxx。算法的Workflow如图xx所示。下面将对主要创新点进行详细介绍。
Building upon this autoregressive framework, we propose FADE. The overall architecture is illustrated in Figure~\ref{fig:workflow}. The subsequent sections will elaborate on the primary innovations within our predictor design.

\begin{figure*}[t]
    \centering
    \includegraphics[width=\textwidth]{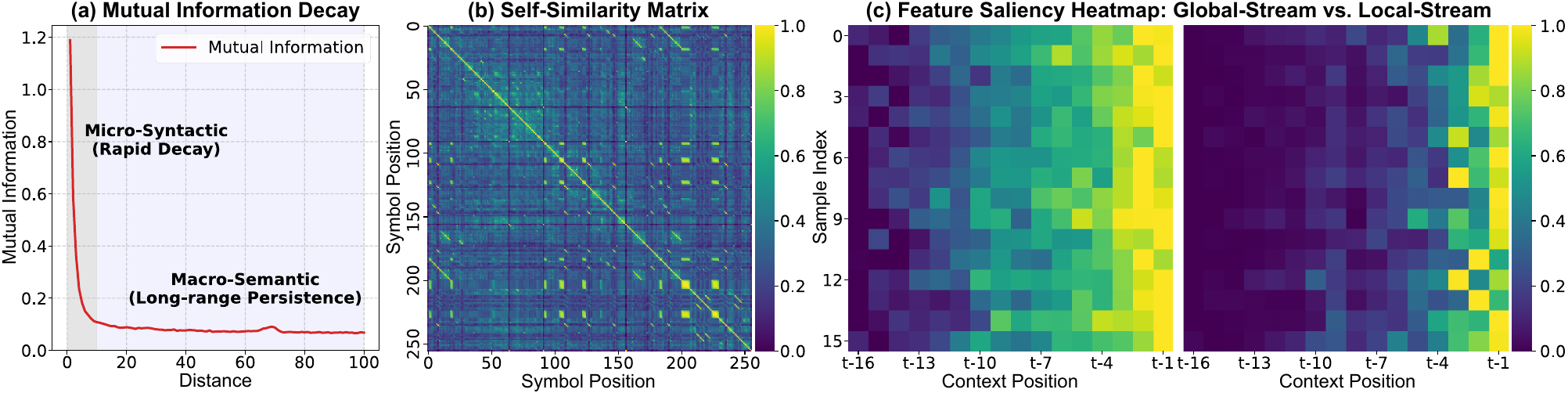}
    \vspace{-0.7cm}
    % \caption{Analysis of dual dependency patterns. (a) and (b) reveal intrinsic micro-syntactic (rapid decay/diagonal) and macro-semantic (long-tail/off-diagonal) features in the data. (c) confirms that our dual-stream architecture effectively decouples these local and global contexts.}
    % \caption{Verification of dual dependency patterns on Silesia. (a) Mutual information decay exhibits a sharp initial drop (micro-syntactic) followed by a persistent non-zero tail (macro-semantic). (b) The self-similarity matrix corroborates this via a prominent diagonal band and recurring off-diagonal blocks. (c) Visualization of our dual-stream architecture effectively disentangling these local and global contexts.}
    \caption{Verification of dual dependency patterns on Silesia. (a) Mutual information decay exhibits a sharp initial drop (micro-syntactic) followed by a persistent non-zero tail (macro-semantic). (b) The self-similarity matrix corroborates this observation via a prominent diagonal band and recurring off-diagonal blocks. (c) Feature saliency heatmaps of the Global and Local streams, illustrating the distinct patterns captured by each branch.}
    \label{fig:heatmap}
    \vspace{-0.3cm}
\end{figure*}

\subsection{Dual-Stream Multi-Scale Decoupler}
\textbf{Analysis}. Information-theoretic studies~\cite{shannon1948mathematical,khandelwal2018sharp} reveal that data sequences exhibit dual dependency patterns: micro-syntactic dependencies governed by local regularities (e.g., N-gram patterns) and macro-semantic dependencies spanning long-range context, empirically verified in Figure~\ref{fig:heatmap}. Existing LDC methods primarily employ MLPs for rapid inference. However, the single-stream MLP inherently functions as a full-scale mixer, attempting to fit these heterogeneous features using a shared set of parameters. As illustrated in Figure~\ref{fig:heatmap} (c), this results in a diffuse saliency distribution that fails to capture sharp micro-syntactic fluctuations, leading to multi-scale interference. To compensate for this lack of specialized inductive bias, existing methods are often compelled to stack deeper layers to approximate complex distributions. While this strategy marginally improves representation, the increased computational depth forces a long sequential execution path, directly translating to higher latency.\\
\textbf{Design.} We propose the Dual-Stream Multi-Scale Decoupler (DMD). By implementing explicit feature decoupling, DMD processes features via parallel streams with distinct inductive biases. Crucially, this design simultaneously compensates for saliency dilution and replaces deep serial stacking with shallow parallel execution. Formally, given the input sequence embedding $\bm{X}_{\text{emb}}\in\mathbb{R}^{B\times T\times D}$ (with batch size $B$, time steps $T$, and embedding dimension $D$) and the flattened normalized input $\bm{X}\in\mathbb{R}^{B\times 1\times D_h}$ (where $D_h=T\times D$), the processing workflow is formulated as follows:\\
\textbf{(1) Global Stream for Macro-scale Modeling}. Dedicated to macro-semantic modeling, this stream employs a GeGLU-based Rolling Cache~\cite{dauphin2017language,shazeer2020glu,PAC,lu2025edpc} to capture long-range dependencies. This design enhances the nonlinear expressivity of historical context while maintaining inference efficiency. Specifically, we maintain a latent cache $\bm{M}\in\mathbb{R}^{B\times 1\times D_{cache}}$, which is updated at step $t$ by integrating the latest feature via a rolling operation:
\begin{equation}
    \text{GeGLU}(\bm{X}_t)=\psi(\bm{X}_t\bm{W}_g)\odot(\bm{X}_t\bm{W}_v)
\end{equation}
\begin{equation}
    \bm{M}_t = \text{Roll}(\bm{M}_{t-1}, \text{GeGLU}(\bm{X}_t))
\end{equation}
where $\bm{X}_t \in \mathbb{R}^{B\times 1 \times D}$ denotes the embedding of the latest symbol (i.e., the last $D$ channels of the input $\bm{X}$), $\odot$ denotes element-wise multiplication, and $\psi$ is the GeLU activation function~\cite{hendrycks2016gaussian}. The updated cache is then projected back into the output space to yield the global feature $\bm{H}_{\text{global}}\in\mathbb{R}^{B\times 1\times D_h}$:
\begin{equation}
    \bm{H}_{\text{global}} = \psi(\bm{M}_t \bm{W}_m) + \lambda_m \cdot \bm{X}
\end{equation}
where $\bm{W}_m\in\mathbb{R}^{D_{\text{cache}}\times D_h}$ denotes the projection matrix, and $\lambda_m$ is a learnable residual scaling factor initialized to 1.\\
\textbf{(2) Local Stream for Micro-scale Modeling}. To address the multi-scale interference, we introduce a lightweight Local Stream serving as a micro-syntactic decoupler. This branch employs a 1D convolution~\cite{bai2018empirical} to impose a strong local inductive bias, yielding the local feature $\bm{H}_{\text{local}} \in\mathbb{R}^{B\times 1\times D_h}$:
\begin{equation}
    \bm{H}_{\text{local}}=\text{Flatten}(\psi(\text{Conv}(\text{LN}(\bm{X}_{\text{emb}}))))
\end{equation}
where $\text{LN}(\cdot)$ denotes Layer Normalization~\cite{ba2016layer}. As illustrated in Figure~\ref{fig:heatmap} (c), this branch exhibits a sharply localized response pattern. It precisely captures micro-syntactic N-gram patterns while filtering out long-range noise, thereby successfully offloading the syntactic matching task from the global stream.\\
\textbf{(3) Content-Adaptive Router}. To achieve dynamic fusion of multi-scale features, we introduce a Content-Adaptive Router. This module generates routing weights $\bm{\alpha}\in\mathbb{R}^{B\times 1\times D_h}$ conditioned on the input context via a matrix $\bm{W}_r\in\mathbb{R}^{D_h\times D_{h}}$:
\begin{equation}
\bm{\alpha} = \sigma(\bm{X}\bm{W}_r)
\end{equation}
where $\sigma$ represents the Sigmoid activation function. The final fused representation $\bm{H}_{\text{mix}}\in\mathbb{R}^{B\times 1\times D_{h}}$ is computed as:
\begin{equation}
\bm{H}_{\text{mix}} = \bm{\alpha} \odot \bm{H}_{\text{global}} + (1-\bm{\alpha}) \odot \bm{H}_{\text{local}}
\end{equation}

\subsection{Hierarchical Gated Refiner}
\textbf{Analysis}. While the DMD effectively integrates multi-scale features along the temporal dimension, its reliance on globally shared parameters limits its adaptability to the non-stationary feature distribution shifts inherent in online compression. In real-world scenarios, instance-specific context exhibits highly heterogeneous channel interaction patterns. Consequently, shared weights fail to achieve deep instance adaptation. Crucially, merely increasing network depth to capture these variations is ineffective; without selective filtering, it amplifies noise, compromising the probability estimation.\\
\textbf{Design.} To tackle these challenges, we introduce the Hierarchical Gated Refiner (HGR). This module adopts a cascaded strategy transitioning from coarse-grained channel interaction to fine-grained nonlinear refinement, facilitating deep instance adaptation by selectively enhancing high-order semantic features while suppressing noise propagation. Formally, given $\bm{H}_{\text{mix}}\in\mathbb{R}^{B\times 1\times D_{h}}$:\\
\textbf{(1) Coarse-grained Channel Interaction.} HGR captures global correlations via Block Matrix Multiplication. Crucially, since we employ online adaptation with stateful batching, the batch index $k$ corresponds to a fixed data stream. Thus, we define $\bm{W}_U \in \mathbb{R}^{B \times d_h \times d_h}$ as a persistent memory, where the $k$-th slice evolves via backpropagation to capture unique patterns. This effectively achieves sample-adaptive channel mixing, thereby tailoring channel interactions to each specific input. We denote the resulting adaptive feature as $\bm{H}_{\text{a}}\in\mathbb{R}^{B\times 1\times D_{\text{h}}}$:
\begin{equation}
\bm{H}_{\text{a}}=\text{Up}(\text{BMM}(\text{Down}(\text{LN}(\bm{H}_{\text{mix}})), \bm{W}_U)),
\end{equation}
where $\text{Down}(\cdot)$ and $\text{Up}(\cdot)$ denote the dimensionality-reducing and expanding projections (mapping between $D_h$ and $d_h$), respectively~\cite{lu2025edpc}. To mitigate noise accumulation from high-order interactions, HGR employs a content-aware self-gating mechanism to selectively suppress irrelevant features, yielding $\bm{H}_{\text{coarse}}$ via a matrix  $\bm{W}_c\in\mathbb{R}^{D_h\times D_{h}}$:
\begin{equation}
    \bm{H}_{\text{coarse}}=(\bm{H}_{\text{a}}\odot \sigma(\bm{H}_{\text{mix}}\bm{W}_c))+\lambda_c\cdot \bm{H}_{\text{mix}},
\end{equation}
\textbf{(2) Fine-grained Nonlinear Refinement.} Building upon the coarse-grained interaction, we further conduct element-wise feature refinement via GeGLU and a projection matrix $\bm{W}_f \in \mathbb{R}^{D_e \times D_h}$, yielding the expanded representation $\bm{H}_{\text{expand}}$ and the final output $\bm{H} \in \mathbb{R}^{B \times 1 \times D_{h}}$:
\begin{equation}
    \bm{H}_{\text{expand}} = \text{GeGLU}(\text{LN}(\bm{H}_{\text{coarse}}))
\end{equation}
\begin{equation}
    \bm{H} = \psi(\text{LN}(\bm{H}_{\text{expand}}\bm{W}_f)) + \lambda_o \cdot \bm{H}_{\text{coarse}}
\end{equation}

\begin{figure}[t]
    \centering
    \includegraphics[width=1\columnwidth]{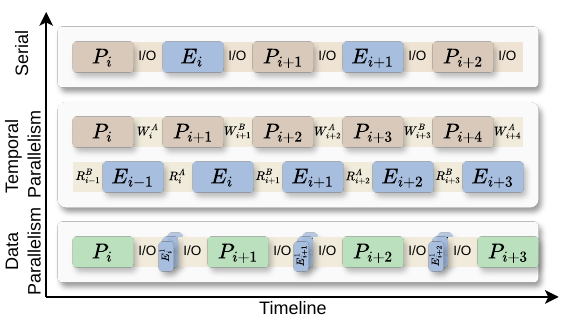}
    % \vspace{-0.6cm}
    \caption{Illustration of execution strategies: Serial, Temporal Parallelism, and Data Parallelism. $P$ and $E$ denote the probability distribution prediction executed on the GPU and the entropy coding process performed on the CPU, respectively. $W^{A/B}$ and $R^{A/B}$ represent writing to and reading from Buffer A/B, respectively.}
    \label{fig:cspp}
    \vspace{-0.35cm}
\end{figure}

\subsection{Concurrent Stream-Parallel Pipeline}
\textbf{Analysis.} 
While advanced pipelines~\cite{wan2025sep,lu2025edpc} mask device heterogeneity, they suffer from a critical limitation: Asymmetry of Parallelism. 
Existing methods accelerate compression but often revert to strictly serial execution during decompression due to autoregressive causality (i.e., $x_t$ depends on $x_{<t}$). 
This results in a performance imbalance where decompression lags significantly behind.\\
\textbf{Design.} 
To address this, we propose the Concurrent Stream-Parallel Pipeline (CSPP), a framework that synchronizes execution strategies across temporal and data dimensions (Figure~\ref{fig:cspp}).\\
\textbf{(1) Temporal Parallelism.} 
To bridge the speed mismatch, we implement an asynchronous pipeline with thread-safe ping-pong buffering. 
This design decouples producer-consumer threads into isolated memory regions. 
Unlike single-buffer queues prone to locking overhead, our zero-copy pointer swapping strategy eliminates memory contention. 
Crucially, this allows the GPU to continuously pre-fetch the next chunk while the CPU processes the current one, masking transmission latency.\\
\textbf{(2) Data Parallelism.}
To resolve the autoregressive bottleneck, we tailor the data parallelism strategy specifically for the autoregressive workflow via a micro-step mechanism.
We partition the input stream into $N$ independent sub-streams to circumvent sequential dependency. 
While each sub-stream maintains internal causality, we orchestrate $N$ workers to execute them concurrently via a dual-barrier protocol.
This effectively transforms the complexity from strictly serial $O(B)$ to parallel $O(B/N)$, boosting overall throughput to match the efficiency of compression.

% \textbf{(2) Data Parallelism}. To maximize parallel hardware occupancy, we shift the optimization perspective to the data dimension by implementing the micro-step Data Parallelism mechanism. We employ stream splitting to partition the batch $B$ among $N$ workers and leverage a dual-barrier protocol to orchestrate execution consistency. Through interleaved barrier coordination of probability and token barriers, the system achieves efficient parallel decoding within micro-step intervals. This fine-grained parallelism reduces the serial proportion along the critical path and lowers complexity from $O(B)$ to $O(B/N)$, effectively breaking the scalability limits of Amdahl's Law to eliminate the decoding bottleneck.

In the compression phase, we integrate both Temporal and Data Parallelism to maximize throughput. In the decompression phase, due to autoregressive causality, we rely exclusively on Data Parallelism.

\begin{table}[t]
\small
\centering
\setlength{\tabcolsep}{4.1pt}
\begin{tabular}{cccc} \toprule
\textbf{Dataset} & \textbf{Type} & \textbf{Size (MB)} & \textbf{Hartley} ($\bm{H}_0$) \\ \hline
Enwik9           & text          & 954                & 7.69             \\
LJSpeech         & audio         & 281                & 8.00             \\
TestImages       & image         & 449                & 8.00             \\
UVG              & video         & 890                & 7.79             \\
CESM             & float         & 954                & 8.00             \\
DNACorpus        & genome        & 654                & 2.00             \\
Silesia          & heterogeneous & 202                & 8.00             \\ \bottomrule
\end{tabular}
\caption{Statistical information of the datasets. $\bm{H}_0=\text{log}\mathcal{|A|}$, where $\mathcal{A}$ represents the alphabet set.}
\label{results-data}
\vspace{-0.5cm}
\end{table}

\section{Experiments}
\label{sec:experiments}

\begin{table*}[b]
\small
\centering
\setlength{\tabcolsep}{5.2pt}
\renewcommand{\arraystretch}{1.2}
\begin{tabular}{cccccccccc} 
\toprule
\multirow{2}{*}{\textbf{Method}} & \multirow{2}{*}{\textbf{Venue}} & \textbf{Enwik9} & \textbf{LJSpeech} & \textbf{TestImages} & \textbf{UVG} & \textbf{CESM} & \textbf{DNACorpus} & \textbf{Silesia} & \multirow{2}{*}{\textbf{Average}$\uparrow$}    \\ 
\cline{3-9}
& & \textbf{text} & \textbf{audio} & \textbf{image} & \textbf{video} & \textbf{float} & \textbf{genome} & \textbf{hete.} &      \\ \hline
\rowcolor{gray-bg}
\multicolumn{10}{c}{\textbf{Traditional Compressor}}   \\
Gzip        & - & 3.100  & 1.168   & 1.359  & 1.578  & 1.369  & 3.685  & 3.133  & 2.199        \\
7z          & -  & 4.689 & 1.370   & 1.670  & 1.887  & 1.829  & 4.450  & 4.352  & 2.892         \\
PBZip2      & -  & 3.936 & 1.363   & 1.723  & 2.054  & 1.413  & 3.805  & 3.878  & 2.596         \\
zstd        & -  & 4.249 & 1.238   & 1.524  & 1.819  & 1.404  & 4.276  & 4.008  & 2.645         \\
\rowcolor{gray-bg}
\multicolumn{10}{c}{\textbf{Learned Compressor}}       \\
DZip   & DCC'21  & 5.758  & 1.257  & 2.146  & 2.456  & 2.488  & 4.448  & 4.661  & 3.316  \\
TRACE  & WWW'22  & 5.142  & 1.783  & 2.290  & 2.336  & 2.696  & 4.278  & 4.517  & 3.292  \\
PAC    & DAC'23  & 5.815  & 1.734  & 2.380  & 2.416  & 2.230  & 4.440  & 4.987  & 3.429  \\
MSDZip & WWW'25  & 5.987  & 1.853  & 2.386  & 2.411  & 2.765  & 4.459  & 5.149  & 3.573  \\
SEP    & IJCAI'25& 6.129  & 1.858  & 2.376  & 2.425  & 2.859  & 4.443  & 5.120  & 3.601  \\
EDPC   & MM'25   & 6.176  & 1.879  & 2.392  & 2.520  & 2.910  & 4.472  & 5.321  & 3.667  \\
FADE (Ours)    & -       & \textbf{6.288}  & \textbf{1.880} & \textbf{2.402} & \textbf{2.603} & \textbf{2.939} & \textbf{4.503} & \textbf{5.400} & \textbf{3.716} \\ \bottomrule
\end{tabular}
\caption{Compression ratios $\uparrow$ of all 11 compressors on 7 datasets. Bold values denote the best results.}
\label{results-cr}
\end{table*}

\subsection{Setup}
\textbf{Dataset.} We employs representative datasets spanning 7 distinct domains, including Enwik9~\cite{Enwik}, LJSpeech~\cite{LJSpeech}, TestImages~\cite{TestImages}, UVG~\cite{mercat2020uvg}, CESM~\cite{zhao2020sdrbench}, DNACorpus~\cite{DNACorpus}, and Silesia~\cite{Silesia}. Details are provided in Table~\ref{results-data}.\\
\textbf{Baselines.} We compare our algorithm with 10 baselines, including 4 classic traditional algorithms: Gzip, 7z~\cite{LZMA}, zstd, and PBZip2~\cite{pbzip2}; and 6 advanced online LDC algorithms: DZip, TRACE, PAC, MSDZip, EDPC, and SEP (based on PAC). Notably, to ensure a fair comparison at the algorithmic level, we forgo the multi-GPU setups used in MSDZip and SEP and instead evaluate all methods on a single GPU using PyTorch's default kernels.\\
\textbf{Metrics.} In this paper, we evaluate performance using Compression Ratio (CR) and Throughput (TP).\\ 
% The CR is defined as $\frac{\text{Size}_{\text{orig.}}}{\text{Size}_{\text{cmp.}}}$, and TP is defined as $\frac{2 \times \text{Size}_{\text{orig.}}}{(\text{Time}_{\text{cmp.}}+\text{Time}_{\text{decmp.}}})$.\\
\textbf{Settings.} To ensure fair comparison, we set the batch size to 512 for all algorithms; all other hyperparameters follow their default settings. Consistent with the advanced method EDPC~\cite{lu2025edpc}, we set the time steps to 16 and the embedding dimension to 32.

All experiments were conducted on a server equipped with an AMD EPYC 7402 24-Core Processor, and 4 $\times$ NVIDIA GeForce RTX 4090 GPUs. The server runs Ubuntu 22.04.5 LTS.

\subsection{Results}
\subsubsection{Compression Ratio}
% We evaluated the CR of all methods across 7 multi-source datasets, with detailed results presented in Table~\ref{results-cr}. Overall, LDC methods yield significantly superior CRs compared to traditional approaches, demonstrating the efficacy of deep neural networks in modeling complex distributions. Among LDC baselines, performance progression aligns with architectural evolution. Specifically, DZip (RNN-based) and TRACE (Performer-based) are limited by long-term dependency modeling or linear attention approximations, respectively. While PAC and MSDZip improve results via ordered masks and cross-layer interactions, and EDPC achieves a strong average CR of 3.667 through multi-branch MLPs, they are ultimately surpassed by our method. FADE achieves the highest Average CR of 3.716, establishing a new state-of-the-art. Attributed to its macro-micro feature decoupling and hierarchical gated refinement, FADE captures multi-scale features more precisely than prior arts. This design yields remarkable gains on complex datasets, reaching 6.288 on Enwik9 and 5.400 on Silesia.
We evaluate the CR of all methods on 7 datasets in Table~\ref{results-cr}. Overall, LDC methods significantly outperform traditional approaches. Among baselines, DZip and TRACE are constrained by limited dependency modeling or attention approximations. While PAC and MSDZip improve performance via masking strategies, and EDPC achieves a strong average CR of 3.667, FADE outperforms these approaches, establishing a new state-of-the-art with an average CR of 3.716. Benefiting from the macro-micro feature decoupling and hierarchical gated refinement, FADE captures multi-scale features more precisely, yielding remarkable gains on Enwik9 (6.288) and Silesia (5.400).

\begin{table*}[t]
\centering
\small
\setlength{\tabcolsep}{5.8pt}
\renewcommand{\arraystretch}{1.2}
\begin{tabular}{cc|cccc|cccc}
\toprule
\multirow{2}{*}{\textbf{Method}} & \textbf{Pipeline} & \multicolumn{3}{c}{\textbf{Throughput (KB/min)$\uparrow$}}& \textbf{Impr.} & \textbf{FLOPs}$\downarrow$ & \textbf{Params}$\downarrow$ & \textbf{Latency}$\downarrow$ & \textbf{PGMU}$\downarrow$  \\
            & \textbf{(Cmp. | Decmp.)} & \textbf{Cmp.}    & \textbf{Decmp.} & \textbf{Total} & \textbf{(\%)}  & \textbf{(G)}   & \textbf{(M)}    & \textbf{(ms)}    & \textbf{(GB)}  \\ \hline
DZip        & Serial | Serial     & 466      & 1365       & 695     & 525.5 & 16.56 & 26.18  & 1.98    & 0.496  \\
TRACE       & Serial | Serial     & 2755     & 2187       & 2438    & 78.3  & 9.13  & \textbf{2.37}   & 1.87    & 0.431  \\
PAC         & Serial | Serial     & 2898     & 2349       & 2595    & 67.5  & \textbf{4.33}  & 8.48   & 1.63    & 0.386 \\
MSDZip      & Serial | Serial     & 1988     & 1814       & 1897    & 129.2 & 6.52  & 12.72  & 2.63    & 0.563 \\
SEP         & Parallel | Serial   & 1954 	 & 1636	      & 1781    & 144.1 & 5.41  & 10.57  & 2.26    & 1.053 \\
EDPC        & Parallel | Serial   & 4391     & 2856       & 3461    & 25.6   & 7.10  & 13.84  & 1.09    & 0.394\\
FADE (Ours)         & Parallel | Parallel         & \textbf{4571} & \textbf{4144} & \textbf{4347} & - & 7.83  & 15.20  & \textbf{1.01}    & \textbf{0.367}    \\ \bottomrule
\end{tabular}
\caption{Efficiencies of LDC methods on Silesia. PGMU represents Peak GPU Memory Usage.}
\vspace{-0.3cm}
\label{tab:efficiency}
\end{table*}

\subsubsection{Throughput}
% Table~\ref{tab:efficiency} compares the throughput of 6 LDC methods. Our FADE achieves the highest total throughput of 4347 KB/min, outperforming the SOTA EDPC by 25.6\%. This performance disparity stems directly from pipeline architectures. Conventional methods (such as DZip, TRACE, and PAC) rely on serial pipelines for both phases, where stop-and-wait overheads cap the maximum speed at 2595 KB/min (PAC). SEP incurs further latency due to additional iterative layers. While EDPC successfully parallelizes compression (4391 KB/min), its decompression remains serial, creating a significant bottleneck (2856 KB/min). In contrast, FADE utilizes the CSPP architecture to achieve full-pipeline parallelism. Crucially, in decompression, FADE breaks serial dependencies via Concurrent Data Parallelism, reducing complexity from $O(B)$ to $O(B/N)$. This boosts decompression throughput to 4144 KB/min (a 45.1\% increase over EDPC), ensuring balanced and maximized system efficiency.
Table~\ref{tab:efficiency} shows that FADE achieves the highest total throughput of 4347 KB/min, outperforming baselines by margins ranging from 25.6\% to 525.5\%. This disparity stems from pipeline architectures: conventional methods (e.g., PAC) are capped by serial stop-and-wait overheads, while EDPC is bottlenecked by serial decompression (2856 KB/min) despite parallel compression. In contrast, FADE utilizes CSPP to achieve full-pipeline parallelism. Crucially, by breaking autoregressive constraints via Concurrent Data Parallelism, FADE reduces decoding complexity from $O(B)$ to $O(B/N)$. This boosts decompression throughput to 4144 KB/min (45.1\% higher than EDPC), ensuring balanced and maximized system efficiency.

\subsubsection{Model Performance}
% We evaluated the computational efficiency of the probability prediction models across all LDC algorithms, with results summarized in Table~\ref{tab:efficiency}. As observed, the RNN-based DZip exhibits the highest FLOPs and parameters. TRACE, utilizing the Performer architecture, achieves fewer parameters but suffers from high FLOPs. Meanwhile, MSDZip and SEP demonstrate relatively high Latency and PGMU, whereas EDPC shows performance comparable to PAC. Notably, although FADE incorporates the DMD and HGR modules which result in higher parameter counts, it achieves the lowest Latency and PGMU. This efficiency stems from its architectural design where the dual branches of the DMD execute in parallel. Furthermore, the overall model relies on fewer but computationally denser modules (i.e., large matrix operations) rather than deeply stacked small blocks. This design minimizes kernel launch overhead, enabling FADE to achieve superior compression rates and faster inference speeds simultaneously.
Table~\ref{tab:efficiency} compares the computational efficiency of all models. The RNN-based DZip exhibits the highest FLOPs and parameter count, while TRACE suffers from high FLOPs. MSDZip and SEP show high Latency and PGMU. Notably, despite higher parameter count due to DMD and HGR, FADE achieves the lowest Latency and PGMU. This efficiency stems from the parallel execution of DMD branches and the use of fewer, computationally denser modules (large matrix operations) rather than deep stacking. This design minimizes kernel launch overhead, enabling superior compression and faster inference simultaneously.

\begin{table}[t]
\centering
\small
\setlength{\tabcolsep}{4.3pt}
\renewcommand{\arraystretch}{1.2}
\begin{tabular}{cccccc}
\toprule
\multicolumn{1}{c}{\multirow{2}{*}{\textbf{Module}}} & \multicolumn{1}{c}{\multirow{2}{*}{\textbf{Base +}}} & \multicolumn{1}{c}{\multirow{2}{*}{\textbf{CR}$\uparrow$}} & \multicolumn{3}{c}{\textbf{TP (KB/min)}$\uparrow$} \\
\cline{4-6}
\multicolumn{1}{c}{} & \multicolumn{1}{c}{}  & \multicolumn{1}{c}{} & \textbf{Cmp.} & \textbf{Decmp.} & \textbf{Total}    \\ \hline
\multirow{2}{*}{DMD} & MLP                   & 3.412              & 6353     & 4853      & 5502     \\
                     & CNN                  & 4.086              & 5600     & 4372      & 4910     \\ \hdashline 
\multirow{3}{*}{HGR} & CCI w/o gate          & 4.408              & 4983     & 3995      & 4435     \\
                     & CCI w/ gate           & 4.565              & 4755     & 3836      & 4246     \\
                     & FNR                   & 5.400              & 3730     & 3188      & 3438     \\ \hdashline 
CSPP                 & CSPP                  & 5.400              & 4571     & 4144      & 4347     \\ \bottomrule
\end{tabular}
\caption{Ablation study on the progressive integration of proposed components on Silesia.}
\label{results-ab}
\vspace{-0.4cm}
\end{table}

\subsection{Ablation Studies}
\subsubsection{Effectiveness of Components}
\label{sec:ablation}
To investigate the efficacy of each component, we conduct a progressive ablation study on the Silesia dataset, starting from the baseline model. Detailed results are provided in Table~\ref{results-ab}. Building on the MLP baseline, DMD utilizes local convolutions to resolve multi-scale interference, elevating the CR from 3.412 to 4.086. Subsequently, the integrated HGR employs coarse-grained channel interaction and fine-grained nonlinear refinement to facilitate deep instance adaptation while effectively suppressing noise propagation, significantly propelling the CR to 5.400. However, the associated computational overhead reduces the TP to 3438 KB/min. Finally, by applying the CSPP we resolve this system bottleneck via fine-grained parallel optimization. This restores the Total TP to 4347 KB/min (representing a substantial 26.4\% gain) while preserving optimal compression performance, achieving a superior balance between efficiency and effectiveness.

\begin{figure}
    \centering
    \includegraphics[width=\columnwidth]{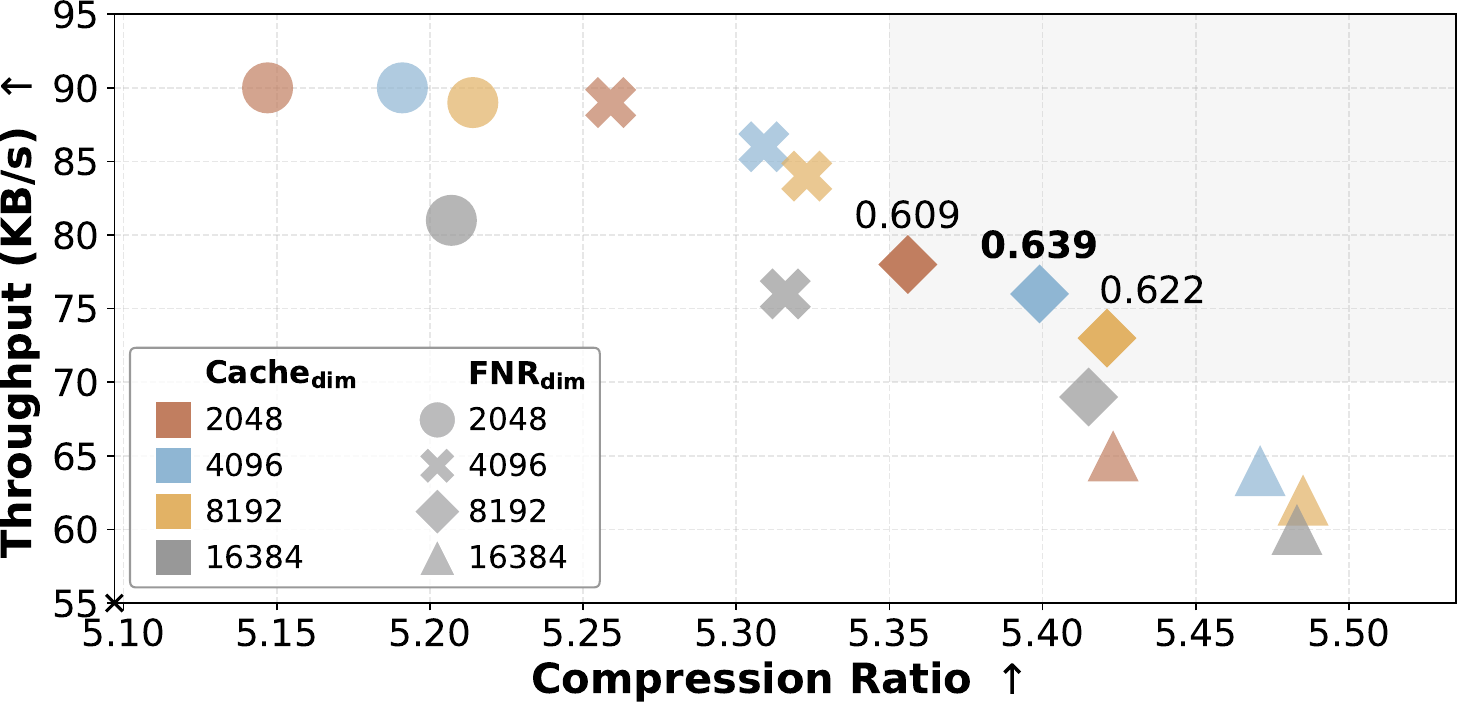}
    \caption{Analysis of different settings on Silesia.}
    \label{fig:dim}
    \vspace{-0.3cm}
\end{figure}

\begin{figure*}[t]
    \centering
    \includegraphics[width=\textwidth]{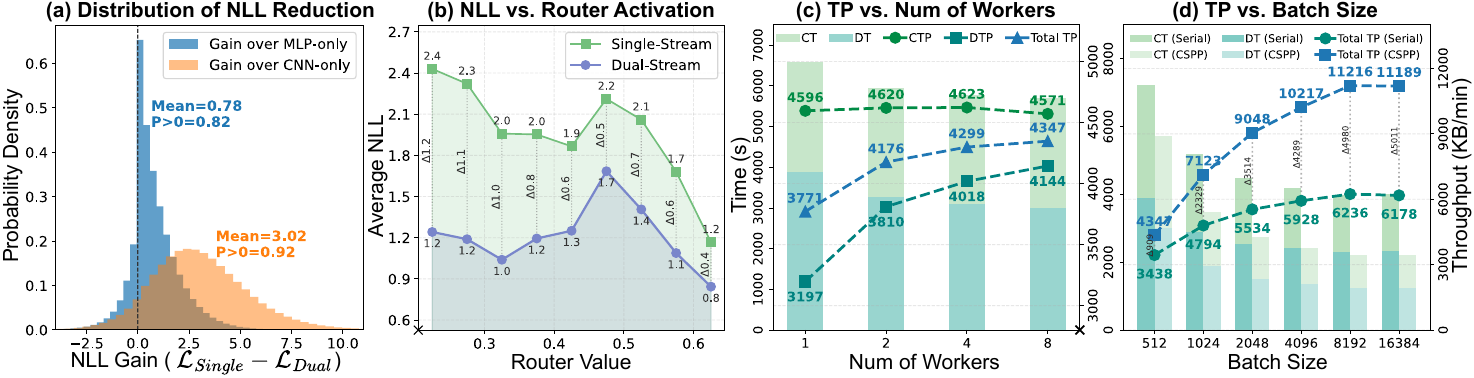}
    \vspace{-0.6cm}
    % \caption{Analysis of NLL characteristics in the dual-stream architecture.}
    \caption{(a-b) NLL characteristics of the dual-stream architecture. (c-d) Impact of worker count and batch size on throughput. CT/DT and CTP/DTP denote Running Time and Throughput of Cmp./Decmp.}
    \label{fig:nll}
    % \vspace{-0.3cm}
\end{figure*}

\subsubsection{Impact of Hidden Dimensions}
We determine optimal model capacity via a grid search on Silesia, varying the hidden dimensions of Rolling Cache ($\text{Cache}_\text{dim}$) and FNR ($\text{FNR}_\text{dim}$) from 2048 to 16384. As shown in Figure~\ref{fig:dim}, larger dimensions improve CR at the cost of latency. To identify the optimal trade-off among Pareto candidates, we calculate the Weighted Normalized Score which is defined as:
\begin{equation}
\scriptsize
\text{Score} = \omega \cdot \frac{\text{CR} - \text{CR}_\text{min}}{\text{CR}_\text{max} - \text{CR}_\text{min}} + (1-\omega) \cdot \frac{\text{TP} - \text{TP}_\text{min}}{\text{TP}_\text{max} - \text{TP}_\text{min}}
\end{equation}
where $\omega=0.5$ balances the metrics, and min/max denote search space extremes. Under constraints of $\text{CR} \ge 5.30$ and $\text{TP} \ge 70$ KB/s, the combination of $\text{Cache}_{\text{dim}}$=4096 and $\text{FNR}_\text{dim}$=8192 yields the highest score (0.639). Consequently, we adopt this configuration as the default for all evaluations.

% \subsubsection{MLP-only vs. Dual-stream}
% To quantify the efficacy of the Dual-Stream Multi-Scale Decoupler, we employ the Negative Log-Likelihood (NLL) to assess distribution fitting capability. The average NLL is defined as 
% \begin{equation}
% \mathcal{L}_{\text{NLL}}=-\frac{1}{T}\sum_{t=1}^{T}\ln P(x_t|x_{<t})
% \end{equation}
% Figure~\ref{fig:nll} (a) illustrates the NLL gains ($\Delta_{\text{NLL}}=\mathcal{L}_{\text{single}}-\mathcal{L}_{\text{dual}}$) of the dual-stream architecture over single-stream baselines. The significant gain over the MLP baseline confirms the success of the local stream in effectively offloading micro-syntactic tasks. Meanwhile, the advantage over the CNN indicates that N-gram features alone are insufficient for modeling complex semantics. Thus, the dual-stream design establishes a benchmark unattainable by isolated architectures. Furthermore, Figure~\ref{fig:nll} (b) reveals the correlation between router activation and prediction difficulty, exhibiting a distinct inverted-U trend. Both ends of the curve correspond to confidence zones with low NLL, demonstrating that the model accurately routes patterns to respective specialized branches. Conversely, peak NLL concentrates in the central ambiguity zone. Notably, the performance gap in the left region validates the decisive role of the local branch in correcting the micro-blind spots of the global model.
\subsubsection{Single-Stream vs. Dual-Stream}
To quantify the efficacy of the Dual-Stream Multi-Scale Decoupler, we employ Negative Log-Likelihood (NLL) to assess distribution fitting capability. The average NLL is defined as
\begin{equation}
\mathcal{L}_{\text{NLL}}=-\frac{1}{T}\sum_{t=1}^{T}\ln P(x_t|x_{<t})
\end{equation}
Figure~\ref{fig:nll} (a) illustrates the NLL gains ($\Delta_{\text{NLL}}=\mathcal{L}_{\text{Single}}-\mathcal{L}_{\text{Dual}}$) of the dual-stream architecture over single-stream baselines. The significant gain over the MLP baseline confirms the local stream effectively offloads micro-syntactic tasks. Meanwhile, the advantage over CNN indicates that N-gram features alone are insufficient for modeling complex semantics. Thus, the dual-stream design establishes a benchmark unattainable by isolated architectures. Furthermore, Figure~\ref{fig:nll} (b) reveals the correlation between router activation and prediction difficulty, exhibiting an inverted-U trend. Both ends correspond to confidence zones with low NLL, demonstrating accurate routing to specialized branches. Conversely, peak NLL concentrates in the central ambiguity zone. Notably, the gap in the left region validates the local branch's decisive role in correcting micro-blind spots of the global model.

\subsubsection{Scalability and Generalizability of CSPP}
We investigate the impact of worker count and batch size on throughput, shown in Figure~\ref{fig:nll} (c) and (d). First, with batch size fixed at 512, increasing workers significantly increases both decompression TP (DTP) and total TP. The growth exhibits diminishing marginal returns, peaking at 4347 KB/min with 8 workers, which verifies the critical role of Data Parallelism in accelerating decoding. Conversely, CTP remains largely insensitive to worker count due to its reliance on Temporal Parallelism. Notably, a slight decline in CTP is observed at 8 workers, attributed to increased scheduling overhead and resource contention from managing threads. Consequently, we adopt 8 workers as the default for FADE. Second, fixing worker count at 8, both serial and parallel total TP increase with batch size. However, the parallel configuration exhibits much steeper growth, widening the performance gap against the serial baseline. Throughput saturates at a batch size of 8192, reaching a peak of 11,216 KB/min. This scalability confirms the efficacy of CSPP in maximizing hardware utilization.

To validate CSPP as a generic framework, we integrated it into baselines. As shown in Table~\ref{tab:add_cspp}, CSPP delivers consistent gains ranging from 14.88\% to 28.38\% across all methods. Notably, it accelerates serial baselines by enabling temporal and data parallelism. Crucially, CSPP even boosts the parallel-optimized EDPC by 23.00\%, proving its portability in resolving residual bottlenecks.

\begin{table}[]
\small
\centering
\setlength{\tabcolsep}{2.6pt}
\renewcommand{\arraystretch}{1.2}
\begin{tabular}{cccccc}
\toprule
\textbf{Pipeline}        & \textbf{TRACE} & \textbf{PAC}   & \textbf{MSDZip} & \textbf{SEP}   & \textbf{EDPC} \\ \hline
Standard        & 2438  & 2595  & 1897   & 1781  & 3461 \\
w/ CSPP (Ours)  & 3130  & 3260  & 2295   & 2046  & 4257 \\
Impr. (\%)	    & 28.38 & 25.63 & 20.98  & 14.88 & 23.00 \\ \bottomrule
\end{tabular}
\caption{Analysis of the CSPP across LDC methods.}
\label{tab:add_cspp}
\vspace{-0.3cm}
\end{table}

% \begin{figure}[t]
%     \centering
%     \includegraphics[width=\columnwidth]{figs/tp.pdf}
%     \caption{Impact of worker count and batch size on Throughput. CT/DT and CTP/DTP represent Running Time and Throughput of Cmp./Decmp., respectively.}
%     \label{fig:tp}
%     % \vspace{-0.4cm}
% \end{figure}

\section{Conclusion}
% 在这部片论文中，我们提出了FADE，实现了sota的压缩效果和吞吐量。我们引入DMD模块实现特征解耦，并通过HGR对特征进行精炼。除此之外，提出CSPP，解决了Serial Processing 的瓶颈大幅度提高吞吐量。实验证明我们相较于baseline实现最优压缩率的同时拥有最高的吞吐量和最低的内存占用。
In this paper, we propose FADE, a general-purpose lossless data compressor that establishes a new state-of-the-art. FADE incorporates the Dual-Stream Multi-Scale Decoupler to decouple features and integrates the Hierarchical Gated Refiner for precise refinement. Furthermore, we propose the Concurrent Stream-Parallel Pipeline, which resolves the serial processing bottleneck and significantly boosts throughput. Experiments demonstrate that FADE achieves superior CR compared to baselines, while simultaneously maintaining the highest throughput and lowest GPU memory usage.

\section*{Limitations}
% 目前，压缩率和吞吐量是数据压缩领域最关注的两个核心指标。FADE的主要设计理念也是优先考虑这两个核心目标。FADE通过特征解耦、精调以及并行流水线，将传统的深层串行架构向浅层并行双流架构的转变，从而导致了FLOPs和Params这两个指标相较于除DZip以外的LDC算法稍有增加，如表xx所示。我们认为这是为了消除串行解码瓶颈而做出的刻意权衡 (deliberate trade-off)。至关重要的是，这种理论上的增加并未阻碍物理性能；正如实验所证明的那样，FADE 依然实现了最低的推理延迟和峰值显存占用（PGMU），成功将并行计算能力转化为现实世界中的速度。
% 在今后，我们会 （优化另外两个目标，通过什么潜在可能的什么方法，帮我写）
Currently, Compression Ratio and Throughput stand as the paramount metrics in modern data compression. The design philosophy of FADE prioritizes these core objectives to meet stringent practical deployment demands. To eliminate the prohibitive serial processing bottleneck, FADE transitions strategically from a conventional deep serial architecture to a shallow parallel dual-stream framework via feature decoupling, hierarchical gated refinement, and a parallel pipeline. This architectural shift results in a marginal increase in FLOPs and parameters compared to LDC baselines other than DZip, as shown in Table~\ref{tab:efficiency}. We consider this a deliberate trade-off necessary to achieve maximal parallelism. Crucially, this theoretical increase in computational cost does not impede real-world efficiency; as evidenced by our experiments, FADE maintains the lowest inference Latency and Peak GPU Memory Usage (PGMU), successfully translating parallel computational capacity into superior speed.

% In future work, we aim to optimize computational complexity (FLOPs) and parameter efficiency (Params) to facilitate lighter-weight deployment. Potential directions include (1) exploring model compression techniques such as network pruning and knowledge distillation to eliminate redundant parameters while maintaining compression performance; and (2) investigating efficient lightweight operators to replace heavy matrix operations, thereby further reducing theoretical computational overhead.

\section*{Acknowledgments}
This work was partly supported by the National Natural Science Foundation of China under Grant (62272252, 62272253) and the China Scholarship Council (CSC) scholarship program.

\balance

\bibliography{custom}

\newpage
\onecolumn
\appendix

\section{Algorithm Description}
The procedure of the online LDC method is outlined in Algorithm~\ref{alg:compression}. The model requires no pre-training; instead, parameters are initialized randomly and updated via backpropagation at each step (Line 9), which is synchronously replicated by the decoder to guarantee lossless reconstruction.
\begin{center} % 让整个块居中
    \begin{minipage}{0.7\linewidth}
        \begin{algorithm}[H]
            \caption{Compression Process of LDC method}
            \label{alg:compression}
            
            \KwIn{Byte Stream $\bm{S}=\{x_i\}_{i=0}^{|\bm{S}|-1}$, Time Step $t$}
            \KwOut{Compressed File $\Phi$}
            
            $P \leftarrow$ Initialize the Probability Predictor\;
            $E \leftarrow$ Initialize the Arithmetic Encoder\;
            
            \For{$i=0$ \KwTo $t-1$}{
                $p(x_i) \leftarrow$ Average probability $\frac{1}{256}$\;
                $\epsilon(x_i) \leftarrow$ Apply $E$ to encode $x_i$ according to $p(x_i)$\;
            }
            
            \For{$i=t$ \KwTo $|\bm{S}|-1$}{
                $p(x_i|x_{i-t},\dots,x_{i-1}) \leftarrow$ Get probability of $x_i$ using $P$\;
                $\epsilon(x_i) \leftarrow$ Apply $E$ to encode $x_i$ according to $p(x_i)$\;
                Backpropagate to update $P$ to minimize the loss\;
            }
            Write binary data $\{\epsilon(x_i)\}_{i=0}^{|\mathcal{S}|-1}$ to the file $\Phi$\;
        \end{algorithm}
    \end{minipage}
\end{center}

\section{Dataset Description}
The detailed descriptions and links of used multi-source datasets are shown in Table~\ref{sup-datainfo}.
\begin{table}[h]
\centering
\small
\setlength{\tabcolsep}{1pt}
\renewcommand{\arraystretch}{1.1}
\begin{tabular}{cccc}
\toprule
Dataset    & Type           & Description                                                                &  Link    \\ \hline
Enwik9     & text           & First $10^9$ bytes of the English Wikipedia dump on 2006.   & \href{https://mattmahoney.net/dc/textdata.html}{Page}   \\
LJSpeech   & audio          & First 10,000 files of the LJSpeech audio dataset.                          & \href{https://keithito.com/LJ-Speech-Dataset/}{Page}     \\
TestImages & image          & A classical 8-bit benchmark dataset for image compression evaluation.      & \href{http://imagecompression.info/test_images/}{Page}     \\
UVG        & video          & The video ShakeNDry from the UVG benchmark featuring 1080p 8-bit YUV format. &\href{https://ultravideo.fi/dataset.html}{Page}  \\
CESM       & float          & First $10^9$ bytes of floating-point data from the CESM-ATM climate dataset. &\href{https://sdrbench.github.io/}{Page}    \\
DNACorpus  & genome         & A corpus of DNA sequences from 15 different species.  &\href{https://sweet.ua.pt/pratas/datasets/DNACorpus.zip}{Page}                          \\
Silesia    & heterogeneous  & A heterogeneous corpus of 12 files covering various file formats.  &\href{https://sun.aei.polsl.pl/~sdeor/index.php?page=silesia}{Page} \\   \bottomrule
\end{tabular}
\caption{Descriptions and links of multi-source datasets.}
\label{sup-datainfo}
\end{table}

\vspace{-0.5cm}

\section{Detailed Information of Baselines}
The implementation details and characteristics of the baselines are show in Table~\ref{sup-methodinfo}.
\begin{table}[h]
\centering
\small
\setlength{\tabcolsep}{10pt}
\renewcommand{\arraystretch}{1.1}
\begin{tabular}{cccccc} \toprule
Method & Ref.  & Version & Language & Methods & Link \\ \hline
\rowcolor{gray-bg}
\multicolumn{6}{c}{\textbf{Traditional Compressor}}   \\
Gzip   & \cite{gzip}  & 1.10 & C/C++    & LZ77, HC        &   \href{https://www.gzip.org/}{Page}  \\
7z     & \cite{LZMA}  & 24.08 & C/C++    & LZ77, AC        &   \href{https://www.7-zip.org/}{Page}   \\
PBZip2 & \cite{pbzip2} & 1.1.13 & C/C++    & BWT, HC         &   \href{https://pyramidresearch.com/pbzip2/}{Page}   \\
zstd   & \cite{zstd}  & 1.5.6 & C/C++    & LZ77, HC        &   \href{https://github.com/facebook/zstd}{Page}   \\
\rowcolor{gray-bg}
\multicolumn{6}{c}{\textbf{Learned Compressor}}   \\
DZip   & \cite{DZip}  & 1.0 & Python   & RNN, AC         &   \href{https://github.com/mohit1997/Dzip-torch}{Page}   \\
TRACE  & \cite{TRACE}  & 1.0 & Python   & Transformer, AC &   \href{https://github.com/mynotwo/A-Fast-Transformer-based-General-Purpose-LosslessCompressor}{Page}   \\
PAC    & \cite{PAC}  & 1.0 & Python   & MLP, AC         &   \href{https://github.com/mynotwo/Faster-and-Stronger-Lossless-Compression-with-Optimized-Autoregressive-Framework}{Page}   \\
MSDZip & \cite{MSDZip}  & 1.0 & Python   & MLP, AC         &   \href{https://github.com/huidong-ma/MSDZip}{Page}   \\
SEP    & \cite{wan2025sep}  & 1.0 & Python   & MLP, AC         &   \href{https://github.com/damonwan1/SEP}{Page}   \\
EDPC   & \cite{lu2025edpc}  & 1.0 & Python   & MLP, AC         &   \href{https://github.com/Magie0/EDPC}{Page}   \\
FADE   & -  & 1.0 & Python   & MLP, CNN, AC         &   \href{https://anonymous.4open.science/r/FADE-D817}{Page} \\ \bottomrule
\end{tabular}
\caption{Implementation details and characteristics of the baseline methods. \textbf{LZ77}: repeated strings are coded by offset and length of previous occurrence; \textbf{HC}: Huffman Coding; \textbf{BWT}: Burrows-Wheeler Transform; \textbf{AC}: Arithmetic Coding.}
\label{sup-methodinfo}
\end{table}

\section{More Experimental Results}
% \begin{wrapfigure}{r}{0.55\textwidth}
%     \centering
%     \includegraphics[width=\linewidth]{figs/loss.pdf}
%     \caption{Model loss trajectories of progressive ablation study.}
%     \label{fig:loss}
%     \vspace{-10pt} 
% \end{wrapfigure}

\subsection{Progressive Ablation Analysis via Loss Trends}
\begin{figure*}[h]
    \centering
    \includegraphics[width=0.8\linewidth]{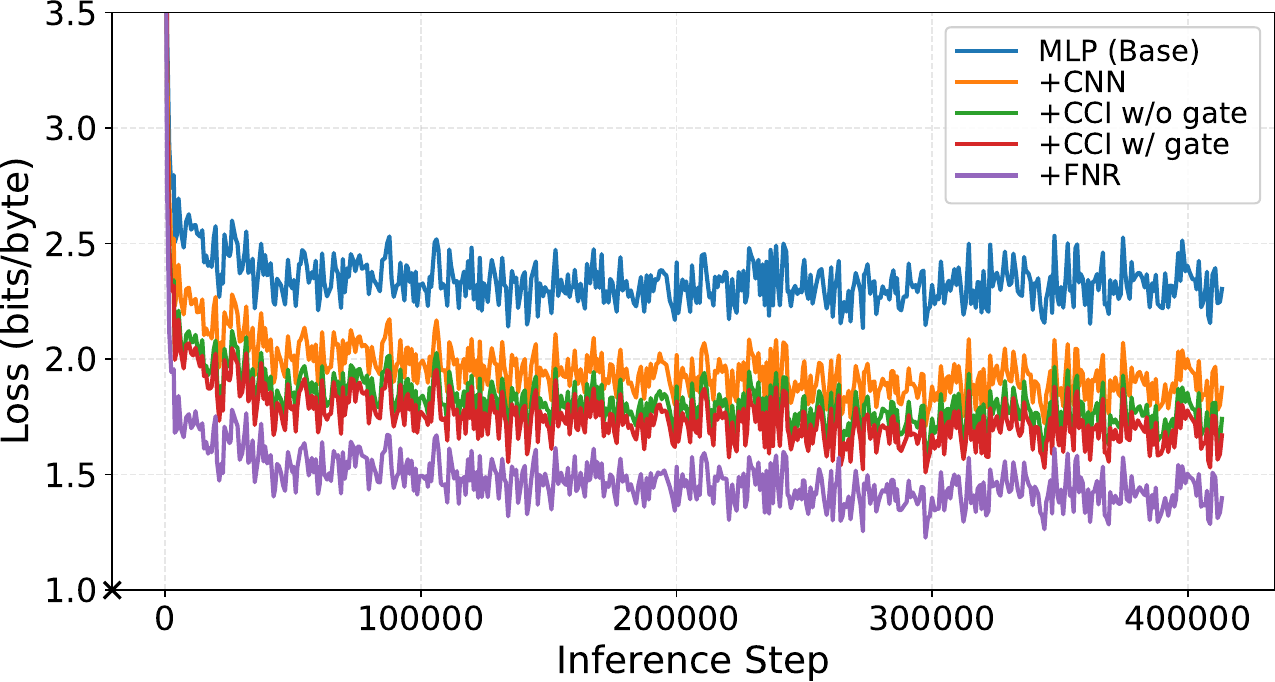}
    \caption{Model loss trajectories of progressive ablation study.}
    \label{fig:loss}
    \vspace{-0.5cm}
\end{figure*}

% In Section~\ref{sec:ablation}, we validated the effectiveness of each component using CR. To provide a more intuitive verification, we further analyze the validation loss trajectories throughout the inference process, as visualized in Figure~\ref{fig:loss}. The pure MLP baseline exhibits the highest entropy and significant volatility, highlighting the instability of relying solely on global features. Introducing the CNN-based local stream triggers a sharp reduction in loss, confirming the validity of the dual-stream decoupling strategy. Furthermore, the gated CCI variant consistently outperforms the non-gated version, proving that the gate effectively filters noise. Finally, the integration of the FNR achieves the lowest loss floor and the most stable convergence trajectory.
In Section~\ref{sec:ablation}, we quantitatively validated the effectiveness of each component using CR. To provide a more intuitive verification, we further analyze the validation loss trajectories throughout the inference process, as visualized in Figure~\ref{fig:loss}. The pure MLP baseline exhibits the highest entropy and significant volatility, highlighting the inherent instability of relying solely on global features. Notably, introducing the CNN-based local stream triggers a sharp reduction in loss, confirming the validity of the dual-stream decoupling strategy. Furthermore, the gated CCI variant consistently outperforms the non-gated version, proving that the gate effectively filters noise. Finally, the integration of the FNR achieves the lowest loss floor and the most stable convergence trajectory.

% \begin{figure}[h]
%     \centering
%     \includegraphics[width=0.7\linewidth]{figs/loss.pdf}
%     \caption{Model loss trajectories of progressive ablation study.}
%     \label{fig:loss}
% \end{figure}

\subsection{Analysis of Content-Adaptive Router}
\begin{figure}[h]
    \centering
    \includegraphics[width=\linewidth]{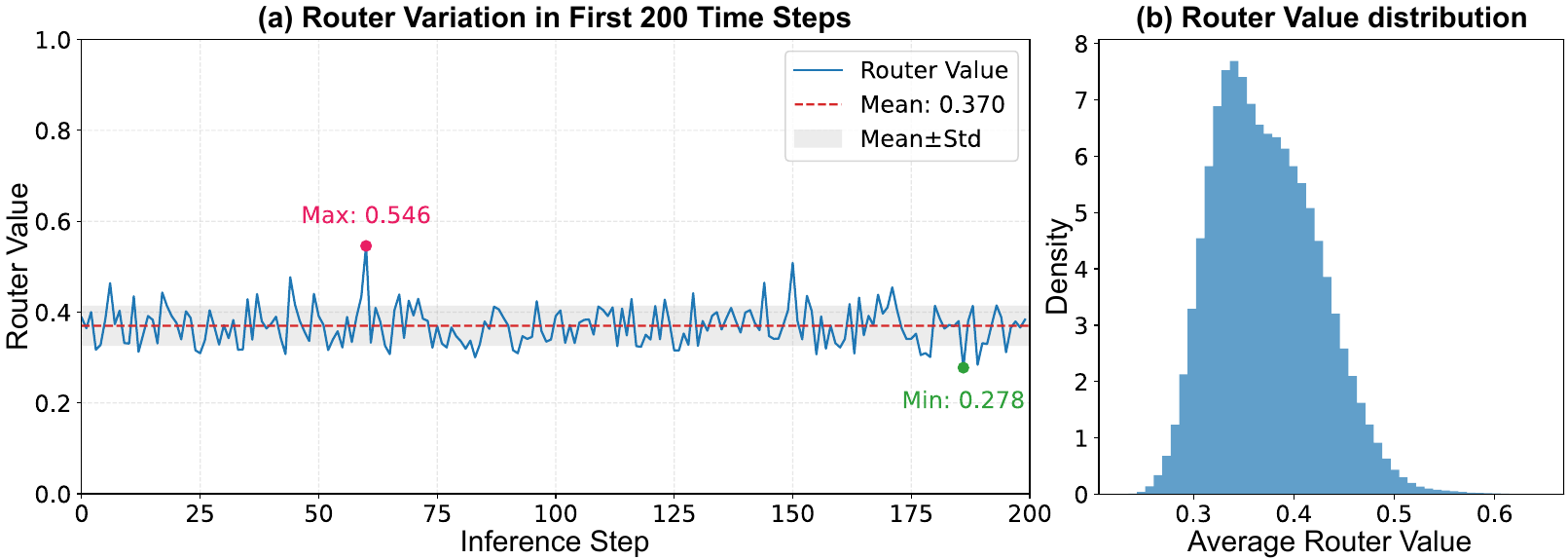}
    \caption{Analysis of the content-adaptive router value dynamics.}
    \label{fig:router}
\end{figure}

To verify the effectiveness of the Content-Adaptive Router, we visualize the dynamic variation of the routing weight $\alpha$ in the first 200 inference steps. As shown in Figure~\ref{fig:router} (a), the router value exhibits high-frequency fluctuations (ranging from 0.278 to 0.546) rather than converging to a static constant. This rapid oscillation confirms the router's sensitivity to micro-contextual changes, allowing it to adjust the fusion strategy symbol-by-symbol.
Consistent with these statistics, the distribution in Figure~\ref{fig:router} (b) displays a unimodal pattern concentrated around this mean value. This indicates a general preference for the Local Stream while retaining the flexibility to incorporate global context when necessary.

\subsection{Impact of Batch Size on CR}
To ensure a fair comparison with existing baselines, we set the default batch size to 512 for the primary evaluation in Section~\ref{sec:experiments}, consistent with their standard settings. Furthermore, we extend our investigation to analyze the impact of batch size on the CR across multi-source datasets, with detailed results presented in Table~\ref{tab:bs_cr}. The results indicate that larger batch sizes generally favor compression performance. Specifically, the model achieves the optimal average CR of 3.755 and 3.754 at batch sizes of 4096 and 8192, respectively. Breaking this down by domain, datasets like Enwik9, UVG, and DNACorpus peak at a batch size of 8192. In contrast, the heterogeneous dataset Silesia achieves its best compression at a smaller batch size of 1024, followed by a gradual decline.

\begin{table}[h]
\centering
\small
\setlength{\tabcolsep}{4pt}
\renewcommand{\arraystretch}{1.3}
\begin{tabular}{ccccccccc|cc} \toprule
\multirow{2}{*}{\textbf{Batch Size}}   & \textbf{Enwik9} & \textbf{LJSpeech}       & \textbf{TestImages}     & \textbf{UVG}    & \textbf{CESM}   & \textbf{DNACorpus}      & \textbf{Silesia }       & \multirow{2}{*}{\textbf{Avg. CR}} & \textbf{FLOPs} & \textbf{PGMU}      \\
\cline{2-8}
 & \textbf{text} & \textbf{audio} & \textbf{image} & \textbf{video} & \textbf{float} & \textbf{genome} & \textbf{hete.} &   & \textbf{(G)} & \textbf{(GB)}   \\ \hline
512 (default)  & 6.288  & 1.880 & 2.402  & 2.603  & 2.939  & 4.503  & 5.400  & 3.716 & 7.83   & 0.367    \\
1024           & 6.365  & 1.884 & 2.407  & 2.613  & 2.952  & 4.515  & 5.409  & 3.735 & 15.65  & 0.509   \\
2048           & 6.423  & 1.888 & 2.410  & 2.623  & 2.951  & 4.548  & 5.390  & 3.748 & 31.31  & 0.796    \\
4096           & 6.465  & 1.888 & 2.411  & 2.631  & 2.954  & 4.566  & 5.371  & 3.755 & 62.61  & 1.375   \\
8192           & 6.491  & 1.887 & 2.409  & 2.643  & 2.948  & 4.568  & 5.335  & 3.754 & 125.22 & 2.532  \\
16384          & 6.486  & 1.883 & 2.404  & 2.639  & 2.936  & 4.561  & 5.268  & 3.740 & 250.44 & 4.847  \\
\bottomrule
\end{tabular}
\caption{Impact of batch size on compression ratio and performance.}
\label{tab:bs_cr}
\end{table}

\begin{figure}[h]
\centering
\includegraphics[width=\linewidth]{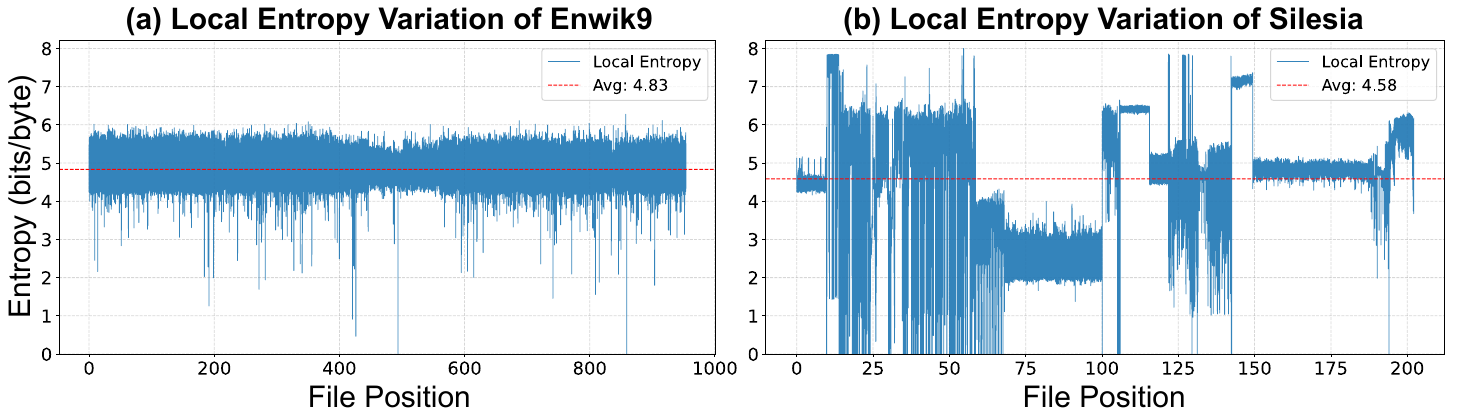}
\caption{Analysis of datasets via local entropy.}
\label{fig:bs_cr}
\end{figure}

To investigate the root cause of this divergence, we visualized the local entropy variations for Enwik9 and Silesia in Figure~\ref{fig:bs_cr}. As observed, Enwik9 exhibits consistent high-frequency fluctuations, indicating a stationary data distribution. Conversely, Silesia displays abrupt jumps and distinct blocky patterns, reflecting its non-stationary and highly heterogeneous nature. This suggests that for stationary sequences, larger batch sizes provide stable global statistics that enhance the HGR's refinement capability. However, for non-stationary data, excessive batch expansion dilutes local distinctiveness, making smaller batches more effective for capturing rapid distribution shifts.

Notably, this scaling benefit is not unbounded. When the batch size increases further to 16,384, the CR degrades across all datasets compared to 8192. This is attributed to context fragmentation, where the cumulative overhead from cold starts outweighs the statistical benefits. Consequently, for practical deployments, we recommend setting the batch size to 4096 or 8192 (contingent on hardware capacity) to achieve an optimal equilibrium between superior compression density and maximum throughput.

\vspace{2em}

\end{document}